\documentclass[10pt,twocolumn,letterpaper]{article}

\usepackage{cvpr}
\usepackage{times}
\usepackage{epsfig}
\usepackage{graphicx}
\usepackage{amsmath}
\usepackage{amssymb}

\usepackage{svg}
\usepackage{epstopdf}
\graphicspath{{./svgs/}}

\cvprfinalcopy % *** Uncomment this line for the final submission

\def\cvprPaperID{5861} % *** Enter the CVPR Paper ID here

\ifcvprfinal\fi
\begin{document}

%%%%%%%%% TITLE
\title{Learning to Infer the Depth Map of a Hand from its Color Image}

\author{Vassilis C. Nicodemou$^{1,2}$\\
{\tt\small nikodim@ics.forth.gr}\\
\and
Iason Oikonomidis$^{1}$\\
{\tt\small oikonom@ics.forth.gr}\\
\and
Georgios Tzimiropoulos$^{3}$\\
{\tt\small yorgos.tzimiropoulos@nottingham.ac.uk}\\
\and
Antonis Argyros$^{1,2}$\\
{\tt\small argyros@ics.forth.gr}\and
$^1$Computational Vision and Robotics Laboratory, Institute of Computer Science, FORTH, Greece\\
$^2$Computer Science Department, University of Crete, Greece\\
$^3$Computer Vision Laboratory, University of Nottingham, United Kingdom
}

\maketitle
%\thispagestyle{empty}

%%%%%%%%% ABSTRACT
\begin{abstract}
We propose the first approach to the problem of inferring the depth map of a human hand based on a single RGB image. We achieve this with a Convolutional Neural Network (CNN) that employs a stacked hourglass model as its main building block. Intermediate supervision is used in several outputs of the proposed architecture in a staged approach. To aid the process of training and inference, hand segmentation masks are also estimated in such an intermediate supervision step, and used to guide the subsequent depth estimation process. In order to train and evaluate the proposed method we compile and make publicly available HandRGBD, a new dataset of 20,601 views of hands, each consisting of an RGB image and an aligned depth map. Based on HandRGBD, we explore variants of the proposed approach in an ablative study and determine the best performing one. The results of an extensive experimental evaluation demonstrate that hand depth estimation from a single RGB frame can be achieved with an accuracy of 22mm, which is comparable to the accuracy achieved by contemporary low-cost depth cameras. Such a 3D reconstruction of hands based on RGB information is valuable as a final result on its own right, but also as an input to several other hand analysis and perception algorithms that require depth input. Essentially, in such a context, the proposed approach bridges the gap between RGB and RGBD, by making all existing RGBD-based methods applicable to RGB input.
\end{abstract}
\interfootnotelinepenalty=10000
%%%%%%%%% TEXT BODY
\section{Introduction}
% INTRODUCTION
\label{sec:introduction}
The task of observing and understanding human activities is of great interest to the field of computer vision. Among other approaches, human activity can be studied by observing and monitoring the state of the human body, either in 2D or in 3D. Particular emphasis is given to the human hands as the interpretation of their behavior is key to understanding the interaction of humans with their environment. Several efforts have been devoted to this direction and important milestones have been achieved~\cite{Sarafianos2016,Yuan2018}.
However, despite the significant progress, a general solution to these problems is still lacking. 

\begin{figure}[t]
\centering
\includegraphics[width=0.48\columnwidth]{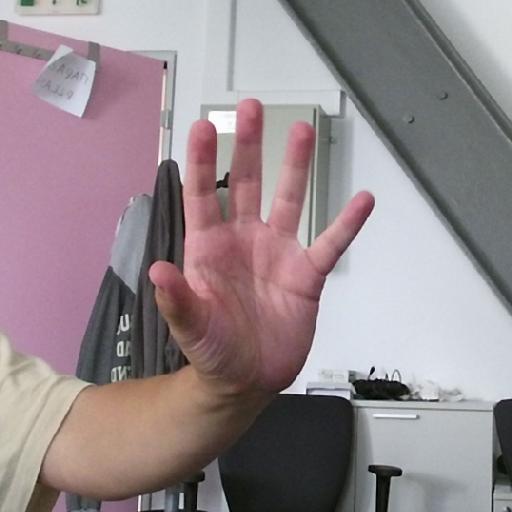}
\includegraphics[width=0.48\columnwidth]{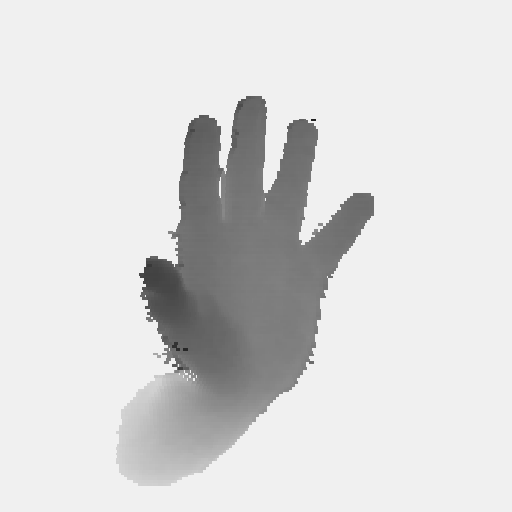}
\caption{Given an RGB image of a human hand (left) the goal of this work is to produce a depth map of the hand region (right).
\label{fig:concept}}
\end{figure}

% WHAT IS THE PROBLEM WE TACKLE?
This work deals with the problem of estimating the depth map of a hand observed from a regular color camera. Depth information is lost during color image formation and is important for the analysis of hands. Our goal is to develop a method that accepts as input a conventional RGB image of a hand and produces the depth map of the observed hand (see Figure~\ref{fig:concept}). The analysis of the rest of the observed scene (non-hand regions) is out of the scope of this work. Thus, no depth  estimation is performed in those regions; the method should only characterize them as background.

% WHAT IS THE SCIENTIFIC/RESEARCH INTEREST?
% WHAT IS THE REAL-WORLD INTEREST? (REAL-WORLD APPLICATIONS)
%
Solving the aforementioned problem is interesting and useful, from both a theoretical and a practical point of view. When observing a scene using regular images, it is very appealing to be able to recover the suppressed depth information without stereo 3D reconstruction or structure from motion. At the same time, the recovery of this information may have significant impact to the solution of several practical problems. As an example, the hand depth information is useful in the context of interaction applications~\cite{Hilliges2012}. On top of this, it can be used to capture and understand hand movement within the 3D space, facilitating tasks such as 3D hand shape and pose estimation, hand-object interaction monitoring, etc, with immediate implications to areas such as human-computer and human-robot interaction, AR/VR, medical rehabilitation, computer games, and more.

% DIFFICULTY OF THE PROBLEM -- WHY ISN'T IT SOLVED?
%
Inferring depth information of hands from color information is a problem that presents important challenges. Human hands exhibit large differences in shape and appearance from person to person. On top of this, due to the articulated structure of the hand, there is a very wide range of 3D postures and  considerable self-occlusions. Further complications arise when the observed hand interacts with its environment, for example when handling objects. Thus, the recovery of the 3D structure of the hand given color information, only, is a rather demanding task.
%
% OTHER APPROACHES (ROUGHLY)
%
Depth estimation techniques using regular color input have been proposed for general scenes~\cite{Torralba2002,Lin2012,Eigen2014} as well as for specific objects such as human faces~\cite{Jackson2017} and bodies~\cite{Varol2018,Pavlakos2018}. However, to the best of our knowledge, no existing method has tackled the problem of hand depth estimation.

% OUR CONTRIBUTION (ROUGHLY)
%
In this work, we capitalize on recent advances in machine learning and propose a deep neural network architecture to tackle the problem of hand depth estimation of hands from color images. 
% Specifically, we adapt a recent architecture~\cite{Jackson2017} that has been successfully applied to the related problem of estimating the 3D structure of human faces.
The training of such a method requires aligned RGB and depth information for a large number of hand views. Up to now, there is no publicly available dataset that is suitable for such a training process. This comes to a surprise, but can be explained by the fact that most of the hand-related works dealt with the problem of 3D hand pose estimation and tracking based on depth information. Therefore, annotation involves depth maps and the associated 3D hand poses and does not include color information. We therefore compiled {\em HandRGBD}\footnote{{\em HandRGBD} will be made publicly available.}, a dataset of $20,601$ pairs of hand color images aligned with their respective depth maps. We use {\em HandRGBD} to evaluate our approach in comparison with variants in the context of an ablation study. We show that hand depth estimation from a single RGB frame can be achieved with an accuracy of $22$mm, which is comparable to the accuracy achieved by contemporary low-cost depth cameras. Thus, the proposed method is a significant step towards turning an RGB camera to an RGBD one for hand analysis applications. Moreover, the proposed method makes all depth-based hand analysis methods exploitable on plain RGB input.

\section{Related Work}

% General-purpose depth from monocular

% \cite{Torralba2002,Saxena2008,Liu2010}

% depth from light field \cite{Chen2014}

% model-based scene sturcture estimation~\cite{Wu2016}

% body depth

% face depth
% ++++

%\vspace*{0.2cm}
{\bf \noindent Depth from color for general scenes:}
The general problem of extracting depth information from color images is very interesting~\cite{Barrow1981,Torralba2002,Saxena2006,Saxena2008,Liu2010}, and is still a research topic under investigation~\cite{Chen2018,Pumarola2018,Mahjourian2018}, remaining unsolved in its full generality.
Since as early as 1985, methods have been proposed for extracting depth information from color images without prior knowledge of the scene~\cite{Barrow1981,Horn1986,Keller1987,Super1995,Shimshoni2000,Torralba2002,Hoiem2005}. The exploited cues vary, including local texture, symmetries, edges and their intersections, and fractal dimension. Also, the extracted level of detail ranges from local depth variations~\cite{Horn1986} to global depth scale of the observed image~\cite{Torralba2002}.
The recent success of machine learning, including (most notably) deep neural networks has naturally led to new methods that achieve increasingly better performance and more accurate results~\cite{Lin2012,Eigen2014,Liu2015,Eigen2015,Laina2016,Garg2016,Li2017,He2018}.
A significant category of methods deals with the problem of estimating the 3D surface of a deformable object~\cite{Varol2009,Salzmann2011}. However, the assumed level of deformation in that line of work is much higher than that of human hands. These methods are designed to tackle deformations of paper and cloth, making the deformation model unnecessarily complicated for the case of human hands. Another category of works on monocular scene structure estimation assumes that the input is a sequence of images, coming from a camera that moves~\cite{Chhaya2016,Mahjourian2018}. Essentially, this results in stereo pairs of images, enabling the use of the disparity cue. A central assumption in this line of work regards the rigidity of the observed world. Due to the articulated structure and motion of the human hand, this assumption does not hold in our setting. Closer to our work, another category of methods uses priors such as the assumption that the scene contains articulated objects~\cite{Wu2016,Wu2018a}. In these approaches, however, the shape and size variation is larger, and the articulation range of the objects is usually more constrained than the case of the human hand.

\vspace*{0.2cm}
{\bf \noindent Depth from color for human body parts:}
There is a recent line of work on methods that tackle the problem of depth estimation for specific parts of the human body. 
The first of these methods~\cite{Jackson2017} estimates the face structure from a single color image. To do so, it uses volumetric information to train a neural network that was based on a stacked hourglass architecture. More works estimating the 3D structure of human faces shortly followed~\cite{Tewari2017,Tewari2018}.
In parallel to the works on human face, a similar architecture was proposed, targeting the human body~\cite{Varol2018}. In that work training data are derived from accurate models of pose, shape and appearance of the human body. None of these approaches achieves a direct estimation of the depth model of the observed scene. Instead, intermediate steps with higher-level information are used, such as the estimation of the landmark positions of facial features for the face approaches, or 2D and 3D position of the body joints in the second case. Moreover, to the best of our knowledge, currently there is no method to solve the problem of estimating depth information from an RGB image of a hand.

\vspace*{0.2cm}
{\bf \noindent Use of depth for 3D hand pose estimation:}
3D hand pose estimation is a long-studied problem~\cite{Rehg1994,heap1996towards,Stenger2001a,Athitsos2003,Sudderth2004,Romero2009,Wang2009,Oikonomidis2011a} that is still of significant interest~\cite{Yuan2018}. Most of the recent works in the area~\cite{Oikonomidis2011a,keskin2012hand,tang2013real,sridhar2013interactive,tang2014latent,oberweger2015hands,tagliasacchi2015robust,wan2016hand,Moon2017,Yuan2018} assume the availability of scene depth information, capitalizing on the advent of inexpensive, high-quality depth sensors. Much more recently, a new trend is currently forming~\cite{zimmermann2017learning,panteleris2017b,Spurr2018,Cai2018,Iqbal2018} that tackles the problem assuming only monocular RGB input. The performance (estimation accuracy and speed) of the older, depth-based approaches is better than that of the more recent RGB-based ones. This is to be expected given the maturity of the older approaches and the richer nature of the depth map as input information.

\vspace*{0.2cm}
{\bf \noindent Our contribution:}
An important goal of this work is to close the gap between depth-based approaches and the newer trend of works based on RGB input. Until today, in order to extract depth information for hands directly, the only available reliable option is to resort to depth sensors. The proposed method estimates hand depth information of comparable accuracy given only regular color images. This constitutes a significant complexity simplification and cost reduction of the sensing process. At the same time, several robust, depth-based hand perception methods become applicable to regular RGB input.
%
% At this point, the distinction between direct depth estimation with input of a color image should be emphasized, as opposed to an indirect estimate which can be achieved by using as an intermediate information, the assessment of the 3D pose and the shape of the observed hand. In the second case, having already highly accurate information about the pose and shape, it is significantly easier to derive a lower level depth information per pixel.
%
In summary, the major contributions of this work are:
\begin{itemize}
\item 
The first method that estimates depth information from monocular color views of hands. This is achieved with an accuracy that is comparable to the accuracy of a low cost depth sensor.
\item 
The {\em HandRGBD} dataset (will become publicly available) of $20,601$ high resolution RGB hand images that are aligned with their depth maps.
\end{itemize}

\section{Hand Depth Estimation from RGB Input}
At the core of the proposed approach, a deep neural network undertakes the task of estimating the geometry of a hand observed in a single RGB image. A stacked hourglass model~\cite{Newell2016} is inspired from parts of the architecture in~\cite{Jackson2017} and used as the main building block for the proposed approach.
The resulting network accepts as input a regular RGB image and outputs the estimated hand depth map. The output of the network is a map of relative depths for all hand pixels of the input image. The absolute depth of the hand is a separate problem~\cite{Torralba2002} that is out of the scope of our work. However, absolute depth estimation can be simplified given a good estimation of the relative depths  resulting from the proposed method.
Intermediate supervision is used in several intermediate levels of the proposed architecture in a staged approach. To aid the process of training and inference, hand segmentation masks are also estimated in such an intermediate supervision step, and used as guidance for the subsequent depth estimation process. 

\subsection{Aligning RGB with Depth}
\label{sec:align}
The training data are assumed to be pairs of aligned RGB and depth hand images. The viewpoint of each image pair is assumed to be identical, i.e., each RGB pixel essentially corresponds to the pixel at the same position in the depth map, as if the two streams were captured from the same center of projection. This kind of data can be obtained using common RGBD sensors like Microsoft Kinect2. Most commercially available RGBD cameras have different sensors for each modality, however the viewpoints are very close, and the availability of depth data enables the alignment of the two streams. To achieve this, an accurate intrinsic and extrinsic calibration of the two sensors is required. Given this, a reprojection of the depth map to the RGB image yields the correspondences between the two images.

\subsubsection{Ground Truth Annotation}
Given the capturing process described in Section~\ref{sec:align}, the training data is already at a usable state and no further manual annotation is required. The only processing that is still required is the segmentation of the RGB and depth channels into foreground (hands) and background (non-hands), and the normalization of the depth range into relative depths. To facilitate this process, it is assumed that the hand is the object closest to the camera. Under this assumption, foreground/background segmentation is easy to perform on the depth map. Towards this end, the minimum value $D_{min}$ in the depth map $D(i,j)$ is estimated, corresponding to the distance of the hands' point that is closest to the camera. The indices $i$ and $j$ run on the horizontal and vertical image dimensions.  
All pixels with a depth value within a predefined threshold $t$ to this minimum depth $D_{min}$ are considered as the foreground $H$. The value $H(i,j)$ of the boolean foreground mask $H$ at point $(i,j)$ is defined as:  
\begin{equation}
    H(i,j) = D(i,j) < (D_{min} + t).
\end{equation}
Working with depth maps in millimeters, it suffices to set $t=300 mm$, a maximum estimation of the possible depth difference within a hand. Since the RGB and depth images correspond pixel-to-pixel, the resulting binary segmentation $H$ is valid for the RGB image, too.

Let us denote with $D[H]$ the depth map $D$ masked with the foreground mask $H$. $D[H]$ is used to compute the average depth value $\overline{D[H]}$ of hand points. 
$\overline{D[H]}$ is then subtracted from $D$ 
and a fixed scaling is applied to the depths, bringing the depth values in the range $[-1,1]$. Specifically, the relative depth map $D_T$ that will be used for training is
\begin{equation}
    D_T(i,j) = c \cdot \left(D(i,j) - \overline{D[H]}\right),
\end{equation}
where $c$ is a value that scales the depths in the range $[-1,1]$. For all cases, it suffices to set $c$ as the inverse of the maximum depth difference of an observed hand. When working with depth values in millimeters it suffices to use $c=1/200$. Finally, the background pixels are set to $1$, the largest value in the target range, that is essentially used to denote background areas.

\begin{figure}[t]
\centering
\def\svgwidth{\columnwidth}
\begin{tiny}
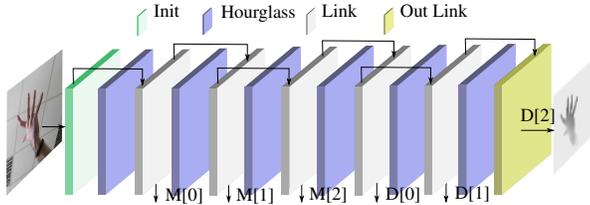
\end{tiny}
\caption{
Stacked Hourglass Architecture: The proposed architecture with the intermediate supervision types. The input is preprocessed by some initialization layers (``Init'', light green) that include a convolutiona layer and two residual blocks (Figure~\ref{fig:convblock} and compute a feature map to be passed to the first hourglass (see Figure~\ref{fig:hourglass}) module. Its output is passed to a set of layers that apply some additional convolution layers (``Link'', gray) before passing it to the next hourglass module. The Link module also outputs a map to be intermediately guided. Skip connections are used parallel to each hourglass module. The first three outputs of the network target segmentation masks and the remaining three target depth maps with the latter being the final output of the network.
\label{fig:ours}}
\end{figure}

\begin{figure}[t]
\centering
%\begin{minipage}[c]{0.45\textwidth}
%\includegraphics[width=\linewidth]{svgs/stacked_landscape.pdf}
%\end{minipage}
%\begin{minipage}[c]{0.49\textwidth}
\def\svgwidth{\columnwidth}
\begin{tiny}
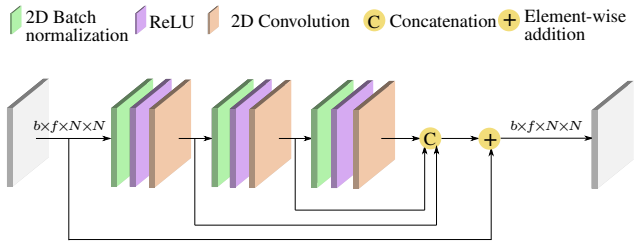
\end{tiny}
%\end{minipage}
\caption{
Residual block: The building block of the proposed neural network for hand depth estimation. The input is assumed to be a feature map of spatial dimension $N \times N$. In the figure, the feature count is $f$. The batch size $b$ is also shown in the tensor dimensions. The output of the block is usually fed to more than one layers, for example to serve a skip connection, as shown here.
\label{fig:convblock}}
\end{figure}

\begin{figure*}[t]
\centering
\def\svgwidth{.9\linewidth}
\begin{tiny}
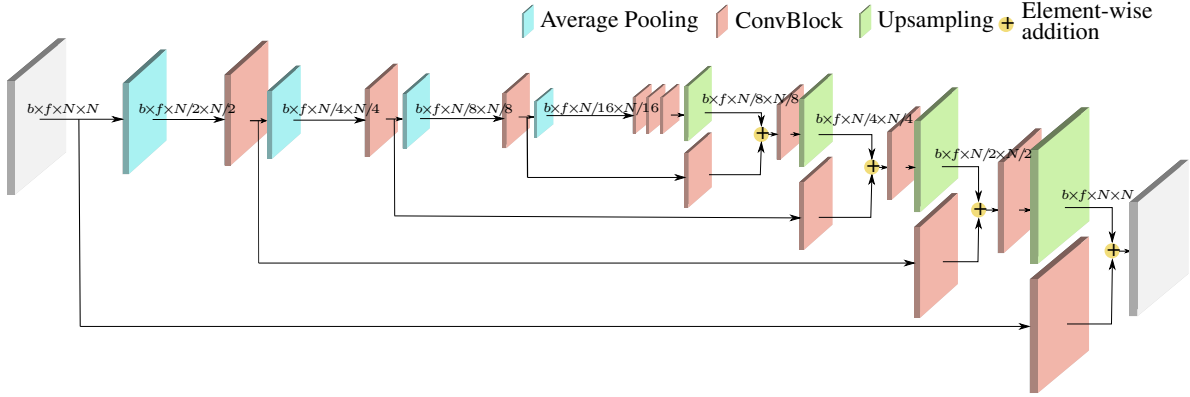
\end{tiny}
\caption{The hourglass building block that is used in the proposed network.
Its main building block is the residual block, illustrated in detail in Figure~\ref{fig:convblock}. The main idea is to successively lower the spatial input resolution for a total of four input halving steps. After this, the reverse procedure is followed to reach again the input resolution.
\label{fig:hourglass}}
\end{figure*}

\subsection{Stacked Hourglass Architecture}
\label{sec:architecture}
The proposed network is based on the approach of stacked hourglass modules~\cite{Newell2016, bulat2017far}. Additionally, intermediate supervision is applied to the output of each hourglass module, which is a commonly adopted strategy~\cite{Newell2016}. The architecture of the proposed method is illustrated in Figure~\ref{fig:ours}. In the following description, the intermediate parts of the network will be called stages.

The main building block of the proposed architecture is the hourglass network of~\cite{Newell2016} built using the residual block of~\cite{bulat2017far}\footnote{Implementation available online at \url{https://github.com/1adrianb/face-alignment}}. Figures~\ref{fig:convblock} and ~\ref{fig:hourglass} illustrate the residual block and the network used graphically.
A hourglass module, illustrated in Figure~\ref{fig:hourglass} accepts as input a set of feature maps. The residual block of~\cite{bulat2017far} proceeds by applying three successive sets of convolution, batch normalization and ReLU non-linearity operations, using also skip connections, similar to the DenseNet architecture~\cite{Huang2017}. This is shown in Figure~\ref{fig:convblock}. After these operations, a down-sampling is performed, halving the input dimension. Parallel to this branch with halved spatial dimension, a skip connection runs through another residual block. In total, four repeated residual blocks and resolution halving are applied, and four long-skip connections run in parallel, each at a different spatial resolution. After the last subsampling and application of a residual block, the reverse process is followed, doubling the spatial dimension by upsampling and applying new residual block operations. After each upsampling, the long-skip connection of the appropriate spatial dimension is added to the current feature map. After four upsampling operations in total, the original input spatial and feature dimension is again reached, forming the complete hourglass module.

For the proposed network, we stack 6 such hourglass modules, having therefore in total 6 stages for intermediate supervision. A convolution operation is applied to the input image to compute a feature map of appropriate dimension to be the input of the first hourglass module. The reverse process is followed at the end of the network, and at the end of each hourglass module for intermediate supervision. Specifically, a single $1 \times 1$ convolution is applied, yielding a single-channel feature map. Each such output is trained against the foreground mask in the first stages, while the later stages are trained against the depth target. We set equal effort for estimating both the mask and the  depth, thus giving $3$ stages for the mask estimation and $3$ for the depth.

% Apart from the target output, the network also produces several intermediate feature maps, one at the end of each stage that comprises a hourglass module. A single convolution on each of these intermediate outputs is applied to yield single-channel outputs. 

% Maybe say something about stages: The architecture uses stages of stacked Hourglass. Each stage has its own output that is the improvement of the previous stages' output. The information that is being carried throughout the stages is the improvement plus any additional estimations, which in our case is the mask that contains the hand.  Because each stage is an improved version of the previous, as it caries all the knowledge and information from the first one, the last stage has the best estimation. 

\subsection{Loss Function}
Each stage has its own target output, therefore each stage has its own loss. The global loss function of the network is the sum of the individual losses. For each stage, regardless of the type of intermediate supervision, its loss is obtained by comparing the two images, the predicted and the target image. We define the loss of each stage as the Mean Square Error (MSE) of the target and the output. The final form of the loss function $L$ is:
\begin{equation}
    L(m,d,\Tilde{m},\Tilde{d}) = \sum_{k=1}^{S_D} \frac{(\Tilde{d}_k-d)^2}{||N||} + \sum_{l=1}^{S_M} \frac{(\Tilde{m}_l-m)^2}{||N||},
    % L(m,x,\Tilde{m_j},\Tilde{x_i}) = \sum_{i=1}^{S_D} \overline{(\Tilde{x_i}-x)^2} + \sum_{j=1}^{S_M} \overline{(\Tilde{m_j}-m)^2}
\end{equation}
where $d$ and $m$ are the target depth and mask, each having $N$ pixels, $S_D$ and $S_M$ are the total number of depth and mask stages respectively and $\Tilde{d}_k$, $\Tilde{m}_l$ are the estimated depths and masks for the $k$th and $l$th stage for $k=1\dots S_D$ and $l=1\dots S_M$.

\subsection{Data Augmentation}
A commonly employed strategy during training is data augmentation which aims at enriching the diversity of the training set and at increasing the generalization capability of the trained network.
% So that the network can learn more generalized scenarios (like different hand illuminations and transformations)
In our case, the input is regular RGB images, and common augmentation practices apply.
Specifically, we apply (a) random horizontal flip (so that we don't have to capture both hands from a subject) (b) random rotation, (c) random crop and (d) random color jittering (to capture the widest possible range of skin tones and different illumination cases). The geometric transformations are applied to both the RGB and the depth maps, ensuring pixel-to-pixel correspondence. The color transformations (e.g., color jittering) is only applied to the RGB channel. Finally, all data are resized to fit the network's input and output dimensions.

\section{Experimental Evaluation}
Due to the unavailability of an appropriate, published dataset, the quantitative evaluation of the proposed hand depth estimation method was performed on the basis of {\em HandRGBD}, a dataset that is introduced in this paper. A first category of experiments assessed the adopted design choices in an ablation study. We also assessed the potential of the proposed method to support methods that perform depth-based estimation of the 3D hand pose.  
% THREE (or four?) types of differences, bg-bg fg-fg bg-fg
%

\subsection{Hand-related Datasets}
Works related to hand appearance and shape modeling as well as pose estimation, require datasets appropriately annotated with ground truth for the purposes of objective, quantitative comparison of competitive approaches, and also - whenever applicable - for training. Therefore, numerous datasets have been proposed so far in the relevant literature (see Table~\ref{tab:datasets}). %~\cite{Gomez-donoso,Simon2017,Bambach2015,Dreuw2006,Yuen2015,tang2014latent,sun2015cascaded,Yuan2017,xu2013efficient,tompson2014real,Tkach2016,zhang20163d,Rogez2014,sridhar2013interactive,Kanhangad2011,zimmermann2017learning,tzionas2015capturing}. 
Input modalities such as monocular RGB, stereo, multiview, and depth are covered. Also, scenarios including egocentric viewpoint, hand-object interaction, and hand-hand interaction are available.

The training and evaluation tasks of the problem we are addressing in this work call for a dataset that includes aligned RGB and depth observations of hands. The RGB input should be unaltered, since the goal is to apply our method to regular color input. Some datasets~\cite{tompson2014real,Tkach2016} warp the RGB image to the depth map, introducing big black holes in the images that defeat this goal. Another dataset~\cite{Kanhangad2011} segments the hand in the image and masks the background with a black color. Given that one of our goals is also to learn this segmentation, the dataset becomes unusable.
Two additional requirements are the presence of multiple actors and close-up views of the depicted hand(s), so that details on the variation of hand shapes across humans and under articulation are adequately captured. 

Table~\ref{tab:datasets} presents a list of the most relevant datasets to our work. Columns of the matrix list some of the requirements listed above, specifically the availability of RGB and depth data, and of their alignment. Evidently, only the datasets by Zimmerman and Brox~\cite{zimmermann2017learning}, called ``Rendered Handpose Dataset'' (RHD) and
Tzionas et al.~\cite{tzionas2015capturing}, called ``Hands in Action'' have aligned RGB and depth data. Unfortunately, the RHD dataset~\cite{zimmermann2017learning} is synthetic, and, although it has a large variation on hand sizes, shapes and appearances, it is of rather low resolution ($320 \times 240$) and contains distant views of a hand. The Hands in Action dataset~\cite{tzionas2015capturing} contains real world data, and the depth is captured by a structured light sensor. The actor diversity is small % prepei na tsekarw an einai mono o dimitris h an exei ki allous
and the view is not closeup, in images of resolution $640 \times 480$. Overall, this dataset comes closest to fulfilling our requirements, however it is still unsuitable due to the somewhat small resolution, the low actor diversity, and (less importantly) the use of a structured light sensor.

% source: https://docs.google.com/spreadsheets/d/1TCjDfm7AmU3TBm00-2UVwvTpFRyYzUHJQeqAK-AE91M/edit?usp=sharing
\begin{table}[t]
    \centering
    \caption{Datasets on human hands. For the purposes of this work, aligned pairs of RGB and depth data are required.}
\begin{tabular}{|l||c|c|c|}
\hline
Dataset & RGB & Depth & Alignment  \\
\hline
\hline
\href{http://www.rovit.ua.es/dataset/mhpdataset/}{Gomez} \cite{Gomez-donoso}                   & \checkmark   & -          & -\\
\hline
\href{http://domedb.perception.cs.cmu.edu/handdb.html}{Simon}  \cite{Simon2017}                & \checkmark   & -          & -\\
\hline
\href{http://vision.soic.indiana.edu/projects/egohands/}{Bambach} \cite{Bambach2015}           & \checkmark   & -          & - \\
\hline
\href{http://www-i6.informatik.rwth-aachen.de/aslr/fingerspelling.php}{Dreuw} \cite{Dreuw2006} & \checkmark   &    -        & - \\
\hline
\href{http://cvrr.ucsd.edu/vivachallenge/index.php/hands/hand-tracking/}{Yuen} \cite{Yuen2015} & \checkmark   &     -       & - \\
\hline
\href{https://labicvl.github.io/hand.html}{Tang} \cite{tang2014latent}                         & -            & \checkmark & - \\
\hline
\href{https://jimmysuen.github.io/}{Sun} \cite{sun2015cascaded}                                & -            & \checkmark & - \\
\hline
\href{http://icvl.ee.ic.ac.uk/hands17/challenge/}{Yuan}  \cite{Yuan2017}                       & -            & \checkmark & - \\
\hline
\href{http://hpes.bii.a-star.edu.sg/}{Xu} \cite{xu2013efficient}                               & -            & \checkmark & - \\
\hline
\href{http://cims.nyu.edu/\%7Etompson/NYU_Hand_Pose_Dataset.htm}{Tompson} \cite{tompson2014real} & Warped     & \checkmark & - \\
\hline
\href{http://lgg.epfl.ch/publications/2016/HModel/index.php} {Tkatch} \cite{Tkach2016}           & Warped     & \checkmark &  -\\
\hline
\href{https://sites.google.com/site/zhjw1988/}{Zhang} \cite{zhang20163d}                         & \checkmark & \checkmark &-  \\
\hline
\href{http://pascal.inrialpes.fr/data2/grogez/UCI-EGO/UCI-EGO.tar.gz}{Rogez} \cite{Rogez2014}    & \checkmark & \checkmark & - \\
\hline
\href{http://handtracker.mpi-inf.mpg.de/projects/handtracker_iccv2013/dexter1.htm}{Sridhar} \cite{sridhar2013interactive} & \checkmark & \checkmark & - \\
\hline
\href{http://www4.comp.polyu.edu.hk/~csajaykr/Database/3Dhand/Hand3DPose.htm}{Kanhangad} \cite{Kanhangad2011} & No BG & \checkmark & \checkmark  \\
\hline
\href{https://lmb.informatik.uni-freiburg.de/resources/datasets/RenderedHandposeDataset.en.html}{Zimmermann} \cite{zimmermann2017learning} & \checkmark & \checkmark & \checkmark  \\
\hline
\href{http://files.is.tue.mpg.de/dtzionas/Hand-Object-Capture/}{Tzionas} \cite{tzionas2015capturing} & \checkmark & \checkmark & \checkmark \\
\hline
\end{tabular}
\label{tab:datasets}
\end{table}

% very useful source for hand datasets (and all hands-related things actually): https://github.com/xinghaochen/awesome-hand-pose-estimation#datasets

\subsection{The {\em HandRGBD} Dataset}
Despite the existence of several hand datasets, it turns out that none of them covers the requirements of this work. Consequently, we resorted to creating {\em HandRGBD}, our own dataset of aligned RGB and depth hand images.

As the capturing device, we employed a Kinect V2~\cite{Kinect2} sensor because of its high quality color camera, and the Time of Flight depth sensor. Among the available options, this sensor provided the best combination of image and depth resolution and quality. The native SDK does not provide an alignment of the depth data to the RGB image, only the opposite, resulting in black holes in the RGB image. Therefore, we used the library libfreenect2~\cite{lingzhu_xiang_2016_50641}\footnote{Source Code available online at \url{https://github.com/OpenKinect/libfreenect2}} that supports this functionality, simultaneously scaling and aligning the depth information on the color image.

The captured dataset contains $20,601$ images along with their respective depth maps. The depicted hands are in closeup view, in distances ranging from $40 cm$ to $100 cm$ from the sensor. Some of the captured images contain two hands that interact (strongly, in some cases). $17$ subjects, $13$ male and $4$ female, contributed to the dataset. The subjects were instructed to keep their hand(s) roughly in the center of the camera field of view, but some images were also captured with hands close to the image edges. The subjects were also instructed to perform free hand gestures and articulations, exploring as much as possible the hand articulation space.
Special care was taken to capture the hands in front of different background scenes, facilitating the generalization of foreground/background segmentation. Also, some of the images contain two hands, that are both annotated as foreground areas.

\subsection{Training Details}
We implemented the proposed approach using the PyTorch framework~\cite{paszke2017automatic}.
The Adam optimizer was used to train it for $100$ epochs, with a learning rate value of $10^{-3}$, weight decay of $10^{-5}$ and a learning rate scheduler with $\gamma=0.5$ applied every $30$ epochs. For training, we employed an Nvidia GTX 1080 Ti GPU. On that machine, each epoch took about $825$ seconds.
For all the experiments, the input size to the network was a $256 \times 256$ RGB image, and the output a $64 \times 64$ depth map.

We split {\em HandRGBD} into training and test sets to train the proposed method. The training set is composed of $19,104$ samples, while the test set contains $1,497$ samples from sequences that are not included in the training set.

% Random Horizontal Flip, Random Rotation, Random Crop, Random Color Jittering
As already mentioned, data augmentation was used in order to increase the generalization of the network. Specifically, each training sample was randomly flipped horizontally with probability $0.5$. Also, a random rotation in the range of $[-90^{\circ}, 90^{\circ}]$ was applied. For the random cropping, a bounding box of size $0.8$ of the original size was selected. Finally, a random intensity value in the range of $[-20, 20]$ for each color channel was added for color jittering.

\subsection{Evaluation Metrics}
{\bf \noindent Assessing depth estimation accuracy:} For each hand pixel we consider the absolute difference between ground truth and estimated depth. The first error metric $E$ (in $mm$) is the average of all these differences for all {\sl actual hand pixels} and all frames of a test set. A second error metric considers the percentage $F(e)$ of hand pixels in the test set for which the absolute difference between ground truth and estimated depth is less that a threshold $e$.

\vspace*{0.2cm}
{\bf \noindent Assessing hand/background segmentation:} The proposed method also produces a segmentation of the hand regions from the background. To assess this, we compute the {\em IoU} (Intesection over Union) criterion for this classification. 

\subsection{Ablative Study}
\label{sec:ablation}
We evaluate different architectural choices  (Section~\ref{sec:architecture}) based on a subset of {\em HandRGBD}. Specifically, variants of the proposed method were trained on $4,500$ images and tested on $500$ separate images of the dataset. 

% \textcolor{red}{Describe here the different variants that were evaluated and comment on the obtained results.}
% We experimented mainly with two parameters of the architecture, namely, the total number of hourglass modules used, and the number of layers with intermediate supervision based on mask.
% Evidently, the models with more hourglass modules perform better. 

An important hyper-parameter of the proposed network is the number of intermediate supervision stages that target the mask segmentation.
Experimenting with different training strategies for the proposed network, it became apparent that the hand segmentation mask is an important cue for the task at hand. In a preliminary experiment, the ground truth segmentation mask was provided as a fourth channel concatenated along with the RGB image to the network. This experiment lowered significantly the depth estimation error, indicating that the segmentation mask is indeed useful. It is therefore important to use this cue as an intermediate supervision target, since it aids the task of the network.

In a network with a fixed number of hourglass modules, some of the first hourglass module outputs target segmentation masks and the rest target depths. We performed an experiment to determine the optimal number of stages for each of the two tasks, experimenting also with the total number of hourglass modules.
The results of this experiment are presented in Table~\ref{table:ablation}. The best results are highlighted with bold font. From this experiment we can conclude that, in fact, the segmentation cue is equally important to the depth map itself. The best performing network with six stages was trained with the first three of them targeted as segmentation mask and the rest targeting depth.

\begin{table}[t]
\caption{Ablative study for the proposed hand depth estimation.}
\begin{center}
\scalebox{0.95}{
\vspace*{-10mm}
\begin{tabular}{|l|c|c|}
\hline 
\cal{Variants of the architecture}  & Error $E$ (mm) & {\em IoU}\tabularnewline
\hline\hline 
\hline 
 0 Mask Stages, 1 Depth Stage    & 39.75 & 0.62 \tabularnewline
% \hline 
% 0 Mask Stages, 2 Depth Stages  & 65.33 & 0.15 \tabularnewline
\hline 
 1 Mask Stage, 2 Depth Stages    & 33.16 & 0.65 \tabularnewline
\hline 
 1 Mask Stage, 3 Depth Stages    & 29.04 & 0.70 \tabularnewline
\hline 
 1 Mask Stage, 4 Depth Stages    & 28.05 & 0.73 \tabularnewline
\hline 
 1 Mask Stage, 5 Depth Stages    & 28.83 & 0.72 \tabularnewline
\hline 
 2 Mask Stages, 4 Depth Stages   & 25.00 & 0.73 \tabularnewline
\hline 
 3 Mask Stages, 3 Depth Stages   & {\bf 24.64} & {\bf 0.81} \tabularnewline
\hline 
 4 Mask Stages, 2 Depth Stages   & 34.42 & 0.68 \tabularnewline
\hline 
\hline 
\end{tabular}}
\label{table:ablation}
\end{center}
\end{table}

\begin{figure}[t]
\centering
\includegraphics[width=\columnwidth]{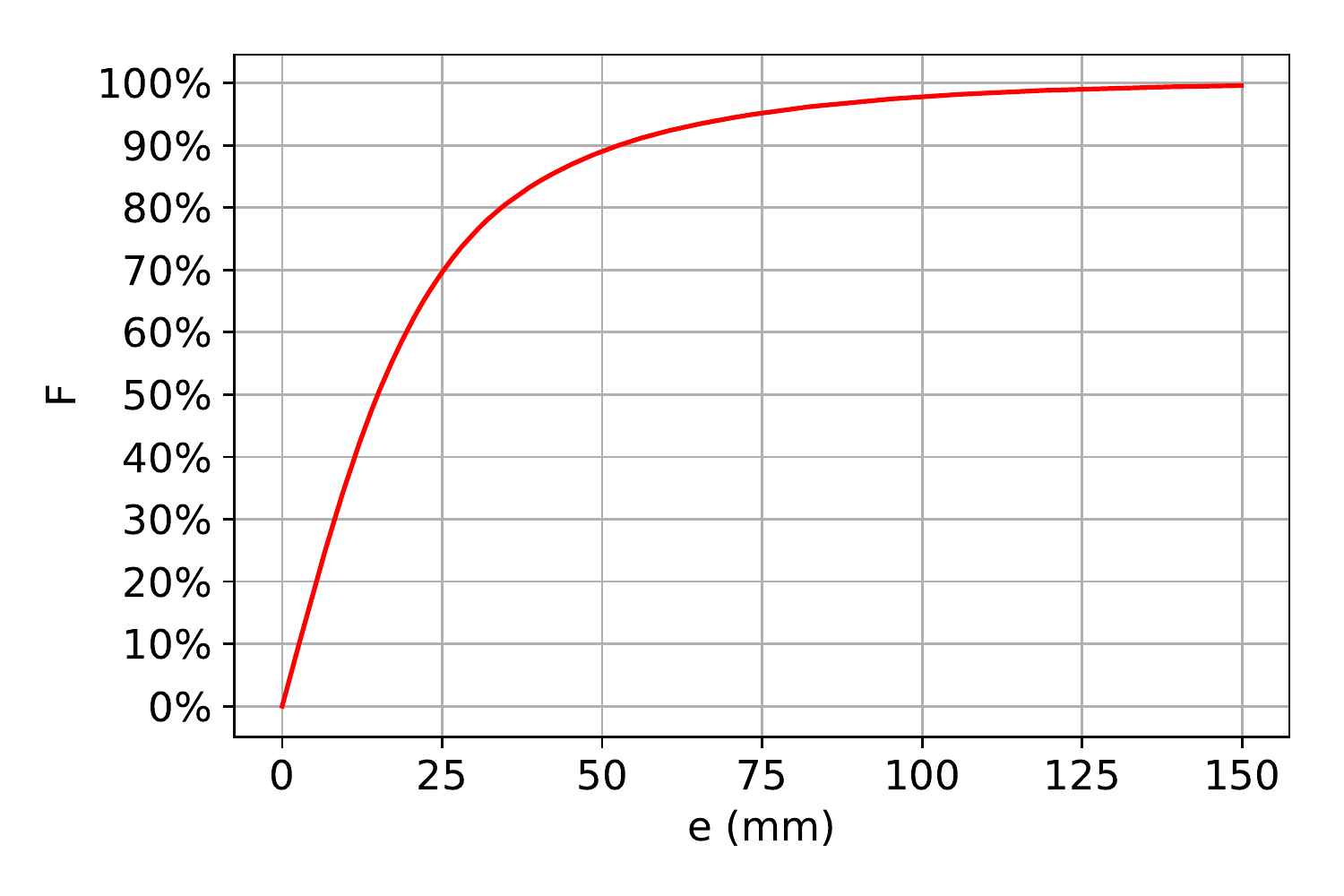}
\caption{The error metric $F(e)$ for the depth accuracy estimation experiment on the {\em HandRGBD} test set (see text for details).
\label{fig:accupercent}}
\end{figure}

\subsection{Hand Depth Estimation Accuracy}
We explored the performance of the best performing variant (line $7$ in Table~\ref{table:ablation}) when trained in a larger subset of {\em HandRGBD}, compared to the experiments in Section~\ref{sec:ablation}. Specifically, we used our training set of $19,104$ images of the dataset for training and the rest $1,497$ images for testing. In this experiment, the depth estimation error $E$ was equal to $E=22.88$mm. Figure~\ref{fig:accupercent} shows the metric $F(e)$. For this experiment, the estimated {\em  IoU} was equal to $0.84$.

\subsection{Supporting 3D Hand Pose Estimation}
We assessed the quality of the depth estimated by the proposed method by evaluating the extend at which it can support depth-based hand pose estimation. To do so, we employed the test set part of {\em HandRGBD} on which we applied a depth-based 3D hand pose estimation method in two different experimental conditions: {\bf C1}, on the actual depth information of the testset as this was measured by the Kinect2 sensor and {\bf C2}, the depth that has been estimated by our method. From the available 3D hand pose estimation methods, we chose to employ the tracking approach of Oikonomidis et al.~\cite{Oikonomidis2011a}. We selected this method because it depends explicitly on the quality of the employed depth map on which it fits a synthetic hand model. This is contrasted to more recent approaches like the one in~\cite{oberweger2015hands}, where hand pose estimation relies on a learned, indirect function of the hand's depth map.

By comparing the performance of~\cite{Oikonomidis2011a} under {\bf C1} and {\bf C2}, we can assess the potential of the proposed method to provide depth maps that are usable by higher level hand perception methods. Ideally, this comparison can be performed by quantifying the 3D hand pose estimation error in {\bf C1} and {\bf C2} based on 3D hand pose ground truth. However, due to the lack of such ground truth, we follow a different strategy. Specifically, we measured the average distance of the corresponding hand joints as those were estimated by~\cite{Oikonomidis2011a} in conditions {\bf C1} and {\bf C2}. This distance was estimated as $24.71mm$. Thus, it turns out that the 3D pose discrepancy between {\bf C1} and {\bf C2} is very similar to the error in depth estimation of our approach on this testset. This is very important, as it provides quantitative evidence that an improvement in RGB-based depth estimation will translate directly to an improvement in 3D hand pose estimation.    

\newcommand{\qualwidth}{0.89\textwidth}
\begin{figure*}[t]
\centering
\includegraphics[width=\qualwidth]{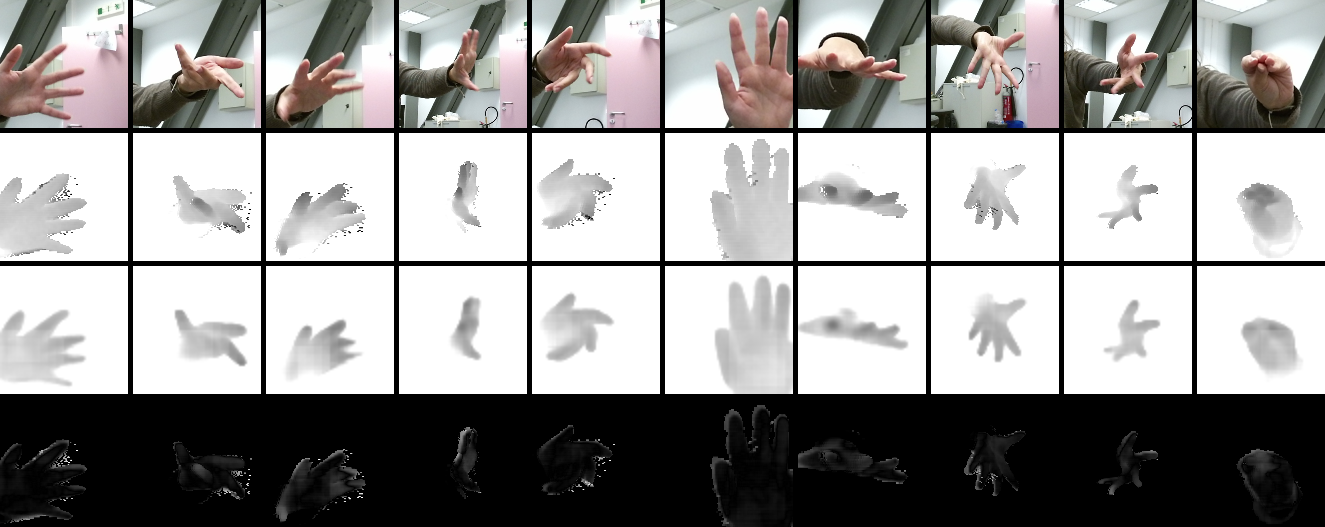}\\ \vspace*{0.1cm}
\includegraphics[width=\qualwidth]{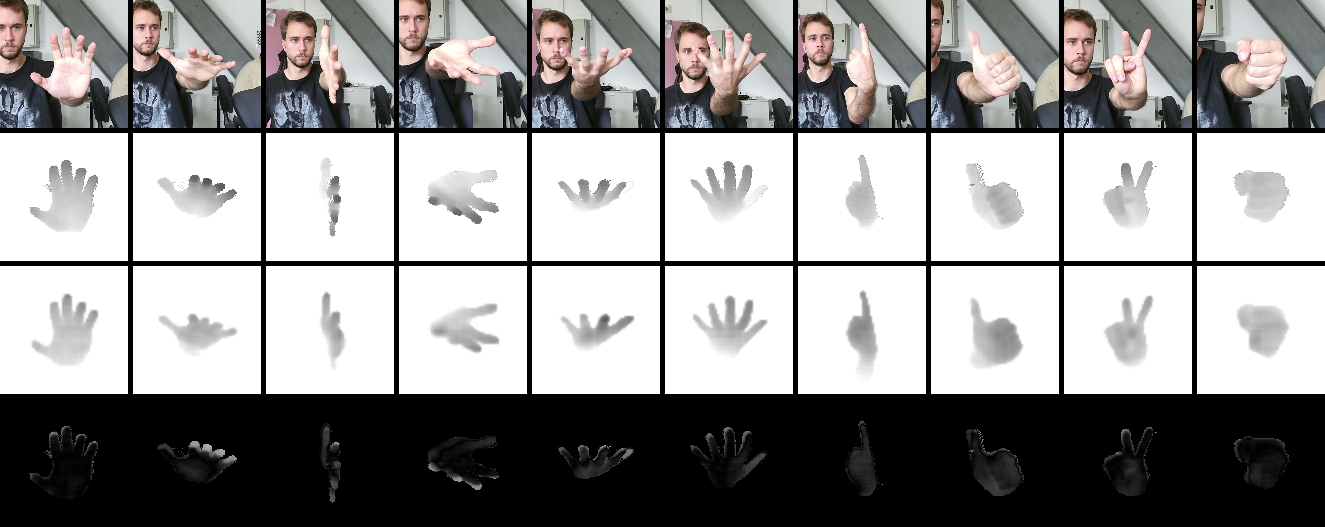}\\ \vspace*{0.1cm}
\includegraphics[width=\qualwidth]{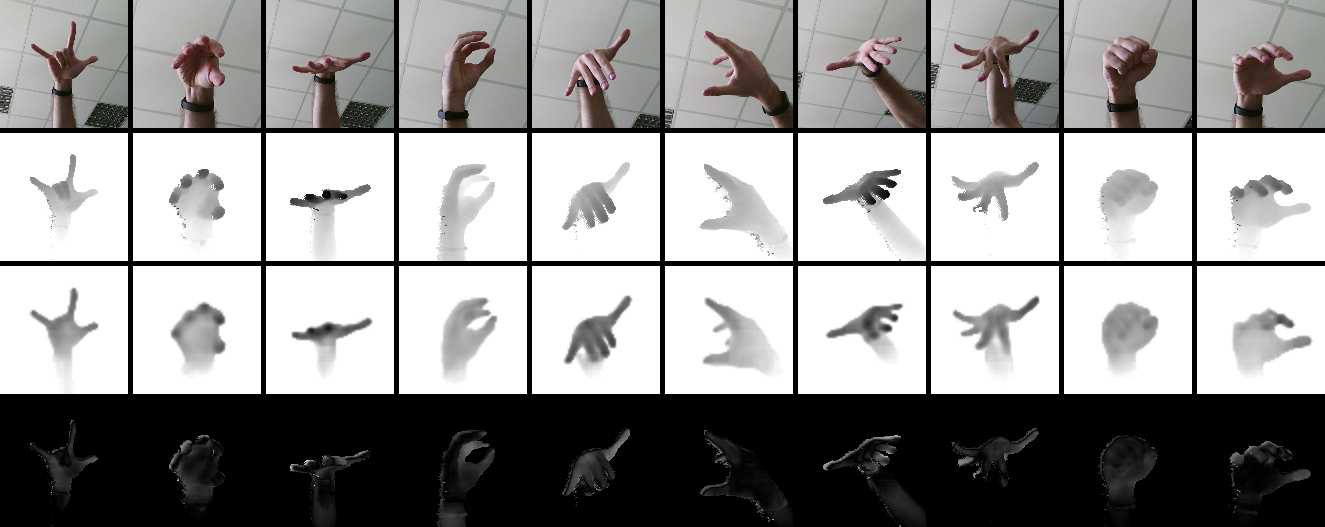}
\caption{Indicative depth estimation results in 3 sequences of {\em HandRGBD}. For each sequence (3 blocks of 4 rows) we show the RGB input (1st row), ground truth depth (2nd row), estimated depth (3rd row) and difference between ground truth and estimated depth (4th row).
\label{fig:qual}}
\end{figure*}

\section{Qualitative Results}
Figure~\ref{fig:qual} shows representative depth estimation results on three sequences of the test set of {\em HandRGBD}. For each sequence, we show the input RGB image, the ground truth depth map, the estimated one, and their color-coded difference. It can be verified that the depth maps estimated by our method are very close to the ones measured by the depth sensor.% More such experimental results are available in the supplementary material accompanying this paper.

\section{Discussion}
We presented the first method that has been specifically designed to estimate the depth map of a human hand based on a single RGB frame. The proposed method consists of a specially designed convolutional neural network that has been trained and evaluated on {\em HandRGBD}, a new dataset of aligned RGB and depth images. Extensive experiments evaluated design choices behind the proposed method, verified its depth estimation accuracy and provided evidence on the potential of the method to support, providing input, existing depth-based hand pose estimation methods. The obtained results demonstrate that for the specific context of hands observation, the proposed method constitutes an important step towards turning a conventional RGB camera to an RGBD one. Future plans include the consideration of the absolute depth estimation problem and the investigation of the suitability of the proposed approach for other RGB-based depth estimation tasks.

{\small
\bibliographystyle{ieee}
\bibliography{egbib,datasets}
}

\end{document}